\documentclass[10pt]{IEEEtran}

\usepackage {graphicx}
\usepackage[cmex10]{amsmath}
\usepackage{cite, algpseudocode, float, url, epsfig, subfigure, stfloats, lipsum}

\begin{document}
\bstctlcite{IEEEexample:BSTcontrol}

\title{Two Dimensional Array Imaging with Beam Steered Data\thanks{This work was in part
supported by the Semiconductor Research Corporation under tasks 1836.054 and 1836.102.}}
\author
{\IEEEauthorblockN{Sujeet Patole and Murat Torlak}\\
\IEEEauthorblockA{Dept. of Electrical Engineering\\
University of Texas at Dallas\\
Richardson, TX 75080, USA}
}
\date{Today}
\maketitle
\begin{abstract}
This paper discusses different approaches used for millimeter wave imaging of two-dimensional objects. Imaging of a two dimensional object requires reflected wave data to be collected across two distinct dimensions. In this paper, we propose a reconstruction method that uses narrowband waveforms along with two dimensional beam steering. The beam is steered in azimuthal and elevation direction, which forms the two distinct dimensions required for the reconstruction. The Reconstruction technique uses inverse Fourier transform along with amplitude and phase correction factors. In addition, this reconstruction technique does not require interpolation of the data in either wavenumber or spatial domain. Use of the two dimensional beam steering offers better performance in the presence of noise compared with the existing methods, such as switched array imaging system. Effects of RF impairments such as quantization of the phase of beam steering weights and timing jitter which add to phase noise, are analyzed.
\end{abstract}

\begin {keywords}
	millimeter wave imaging, beamforming, antenna array, phase noise, two dimensional image reconstruction, Fourier transforms, MATLAB
\end {keywords}

\section{Introduction}
Several approaches are followed for collecting the reflected wave data from the two dimensional target. For example, the reflected wave data can be collected over one-dimensional aperture by simultaneously illuminating the two dimensional target using all antenna elements and collecting response as sum of individual reflected signals \cite{soumekh}. Due to geometry of the array, the reflected wave data arrive at different antenna locations at different time. This fact is used to steer the beam in particular direction. With the one dimensional array the beam is steered with respect to broadside direction forming one of the distinct dimension. To form another distinct dimension required for reconstruction, wideband waveforms are used. Band of frequencies can be covered by the means of FM-CW pulses. Instead of using the electronic beam steering, data can also be collected by rotating an object or radar over different directions. In this case, using the wideband waveforms and ISAR techniques target image is reconstructed \cite{odzemir1}.

In another method, which uses the narrowband waveforms, data is collected using two dimensional antenna array. Only one antenna element in the array illuminates the target at a given time. Receiver antenna element located approximately at the same position as the transmitting antenna collects the reflected wave data from the target. This response is noted along with the position of antenna elements in action. Same process is repeated for each antenna element over the two dimensional aperture providing the reflected wave data collected along the two distinct dimensions. Further in the paper, this method is referred as "switched array method" \cite{sheen}.

This paper introduces new method, which uses the narrowband waveforms along with the beam steering in two dimensions. The beam is steered in both azimuthal and elevation directions, using the two dimensional antenna array. All the antenna elements radiate simultaneously and the waveforms are transmitted using the proper phase shifts at the each antenna so that the combined beam is steered in both the azimuth and elevation directions, covering the complete target. Same process is followed for collecting the reflected wave data across the two distinct dimensions. Reconstruction is carried out using the amplitude and phase correction factors along with the inverse Fourier transform technique.

The method is derived on the similar lines of the 	previous work in this field. However, the reconstruction algorithm is much simple. The beam steering angles in our method are chosen such that the reconstruction process does not require interpolation in wavenumber or spatial domain \cite{soumekh1}.

Resolution of the reconstructed image is function of several parameters such as the antenna array size, spacing between them and the frequency of imaging system. The frequency of the imaging system puts higher limit on the resolution obtained. In order to have the higher resolution, attempts are made to realize millimeter and sub-millimeter wave circuits on CMOS \cite{o}.  Also as the aperture size increases, better resolution can be obtained. The aperture size is a function of the number of antenna elements and the spacing between them. The reconstructed target image quality also depends on focus distance, which is function of how much far the target is placed from the imaging system.

Compared with the switched array method, the two dimensional beam steering method of reconstructing the target performs better in the presence of additive white Gaussian noise. The gain is obtained due to the transmitter scanning the target with the powerful beam formed using beam steering and the then the receiver coherently combining the signals coming from the particular direction. In other words, gain is obtained due to both transmit and receive beamforming. Again, the beamforming gain depends on the number of antenna elements used.

RF impairments such as quantization of the phase of beam steering weights and timing jitter while sampling add to the phase noise. This results into increase in sidelobes of the steered beam. The sidelobes increase with the increase in variance of phase noise. However, this behavior also depends on the aperture size and is characterized in this paper using numerical analysis.

In the paper, section \ref{system_model} explains the system model and notations used throughout the paper. Section \ref{sym_eqn} derives the mathematical expression for reconstructed image. Section \ref{sym_para} discusses effect of various system parameters on the resolution of the system. Section \ref{sect_switched} reviews switched array method, derives reconstruction equation using matched filtering and compares its performance in the presence of noise with the proposed method. Impacts of phase noise on image reconstruction are analyzed in more detail in section \ref{phasenoise}. Simulation results are given in section \ref{symres}, followed by the conclusion.

\section{System Model}
\label{system_model}
As shown in Fig.~\ref {fig_symgeom} the imaging system is located in $X-Y$ plane. Coordinates of a  transmitter antenna are identified by $(a,b)$ and those of a receiver antenna by $(u,v)$. Length of the aperture is limited to  $(-L,L)$ in both transmit and receive mode. Size of the two dimensional array is $ N_a\times N_a$ where $N_a$ is the number of antenna elements per dimension. $\triangle_{xy}$ is the antenna spacing. Thus, $\frac{N_a-1}{2}\times \triangle_{xy}=L$ is the maximum dimension of the aperture. $\lambda$ is the wavelength of the imaging system and $k=2\pi/\lambda$ is the corresponding wavenumber. The target is placed at the distance $z_0$ from the imaging system and its location is identified by Cartesian coordinates $(x,y,z_0 )$.  Two-dimensional target is characterized by its reflectivity function $f (x,y)$. Our aim is to recover this reflectivity function from the reflected wave data. The target is illuminated with the directional beam formed using a two dimensional antenna array. With the antenna array in $X-Y$ plane and the target located in $z=z_0$ plane, the transmitted and received beams makes an angle  $\phi$ in azimuthal direction and $\theta$ in the elevation direction. While recording the reflected wave data, the $(\theta,\phi)$ are varied such that the complete target is covered.	

\section{System Equations}
\label{sym_eqn}
\begin{figure}
\centering
\includegraphics[width=3.2in]{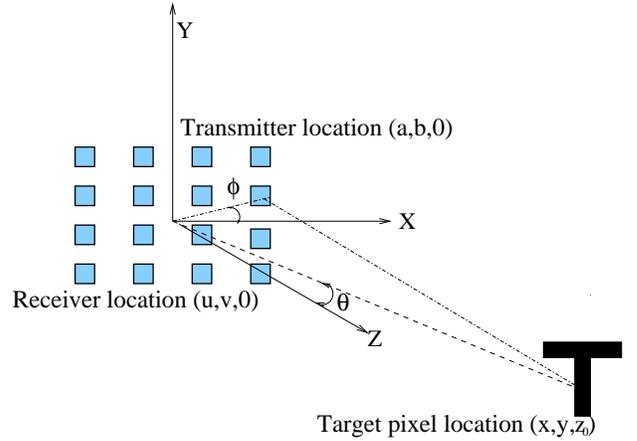}
\caption{Beam steering method}
\label{fig_symgeom}
\end{figure}
\label{sym_eqn}
\begin{figure}
\centering
\includegraphics[width=3.2in]{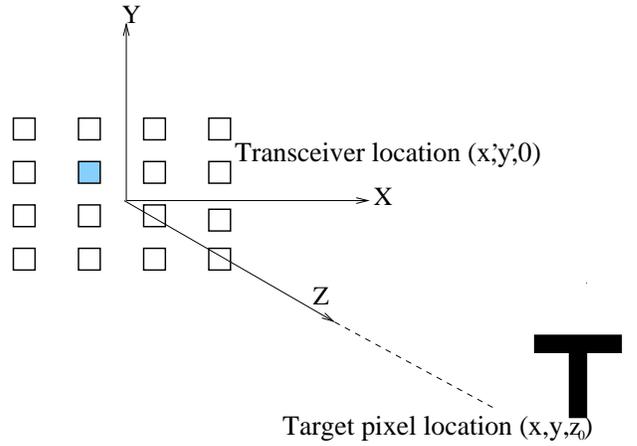}
\caption{Switched array method}
\label{fig_switch}
\end{figure}
As illustrated in Fig.~\ref{fig_symgeom} for the transmitting and receiving antenna elements are placed on the rectangular grid. Using the far field assumption, for a plane wave arriving at an angle $(\theta,\phi)$, delay between instances of arrival between the receiver located at $(a,b)$ and the receiver located at the origin is given by\cite{trees}
\begin{equation}
\label{eqn_timedelay}
\Delta t=\frac{(a\sin\theta\cos\phi + b\sin\theta\sin\phi)}{c}
\end{equation}
thus, the resulting phase difference is given by,
\begin{equation}
\label{eqn_phasediff}
\Delta \Phi=k(a\sin\theta\cos\phi + b\sin\theta\sin\phi).
\end{equation}
Hence, in order to steer the transmitted beam in the direction $(\theta,\phi)$ this phase shift due to geometry should be compensated. This is done by multiplying the signal at the each transmitting antenna by a complex weight that cancels the phase shift. Thus, the resultant signal is added coherently in only one particular direction$(\theta,\phi)$. For example, to steer beam in the direction $(\theta,\phi)$ the complex weight would be,
\begin{equation}
\label{eqn_weights}
W_{\theta\phi}(a,b)=\exp(-jk(a\sin\theta\cos\phi + b\sin\theta\sin\phi)).
\end{equation}
At the target location $(x,y,z_0)$ the signal incident after the transmit beam-steering would be summation of the individual transmitting antenna waveforms scaled by phase shift due to travel delay,
\begin{IEEEeqnarray}{rCl}
\label{Txbeam}
T_{xy}(\theta,\phi)&=&\int\int\exp(-jk\sqrt{(x-a)^2+(y-b)^2+z_0^2})\nonumber
\\ &&W_{\theta\phi}(a,b)d_ad_b.
\end{IEEEeqnarray}
This signal gets reflected back from the target point $(x,y,z_0)$ and then received by the antenna located at $(u,v)$. The receiver multiplies received waveforms at each antenna by the same beam steering weights $W_{\theta\phi}(u,v)$, so that the signals from only particular direction $(\theta,\phi)$ are received. Thus, the total received signal from the target $(x,y,z_0)$ in the particular direction $(\theta,\phi)$ is given by,
\begin{IEEEeqnarray}{rCl}
\label{Rxsig}
s_{xy}(\theta,\phi)&=&\int\int\exp(-jk\sqrt{(x-u)^2+(y-v)^2+z_0^2})\nonumber
\\ &&W_{\theta\phi}(u,v)T_{xy}(\theta,\phi)d_ud_v.
\end{IEEEeqnarray}
The term $exp(-jk\sqrt{(x-u)^2+(y-v)^2+z_0^2})$ in (\ref{Rxsig}) indicates the spherical wavefront originating from point source at $(x,y,z_0)$, which can be decomposed using Fourier transform technique into  superposition of plane waves,
\setlength{\arraycolsep}{0.0em}
\begin{eqnarray}
\label{Fourier_decomp}
&&\exp(-jk\sqrt{(x-u)^2+(y-v)^2+z_0^2})=\int\int R(k_u,k_v)\nonumber\\
&&\exp(\!\!-j\!(\!k_u(x\!-\!u)\!+\!k_v(y\!-\!v)\!+\!\!\sqrt{k^2-k_u^2-k_v^2}z_0\!)\!)dk_udk_v
\end{eqnarray}
\setlength{\arraycolsep}{5pt}
where, $R(k_u,k_v)=\frac{1}{\sqrt{k^2-k_u^2-k_v^2}}$ for omnidirectional case. Making use of this decomposition the received signal $s_{xy}(\theta,\phi)$ can be rewritten as,
\setlength{\arraycolsep}{0.0em}
\begin{eqnarray}
\label{tot_rx}
&&s_{xy}(\theta,\phi)=\nonumber\\
&&\int\int R(k_u,k_v)exp(-j(k_ux+k_vy+\sqrt{k^2-k_u^2-k_v^2}z_0))\nonumber\\
&&dk_udk_v\int\exp(j(k_u-k\sin\theta\cos\phi)du\nonumber\\
&&\int\exp(j(k_v-k\sin\theta\sin\phi)dv\nonumber\\
&&\int\int R(k_a,k_b)exp(-j(k_ax+k_by+\sqrt{k^2-k_a^2-k_b^2}z_0))\nonumber\\
&&dk_adk_b\int\exp(j(k_a-k\sin\theta\cos\phi)da\nonumber\\
&&\int\exp(j(k_b-k\sin\theta\sin\phi)db.
\end{eqnarray}
\setlength{\arraycolsep}{5pt}
Considering infinite length of aperture, all the line integrals can be reduced in following fashion,
\begin{equation}
\label{eqn_lineint}
\int exp(j(k_u-k\sin\theta\cos\phi)du=\delta(j(k_u-k\sin\theta\cos\phi).
\end{equation}
Thus, substituting $k_u, k_a=k\sin\theta\cos\phi$ and $k_v, k_b=k\sin\theta\sin\phi$ (\ref{tot_rx}) can be rewritten as,
\setlength{\arraycolsep}{0.0em}
\begin{eqnarray}
\label{ptresp}
&&s_{xy}(\theta,\phi)=R^2(k\sin\theta\cos\phi, k\sin\theta\sin\phi)\nonumber\\
&&\exp(-j2k(x\sin\theta\cos\phi+y\sin\theta\sin\phi+z_0\cos\theta)).
\end{eqnarray}
\setlength{\arraycolsep}{5pt}
The term $s_{xy}$ in (\ref{ptresp}) denotes response of the point target located at $(x,y)$ with $f(x,y)=1$. Total received signal is superposition of responses from all the target points scaled by respective reflectivity function $f(x,y)$:
\setlength{\arraycolsep}{0.0em}
\begin{eqnarray}
\label{target_resp}
&&s(\theta,\phi)=R^2(k\sin\theta\cos\phi,k\sin\theta\sin\phi)\!\!\int\!\!\int\! \!\!f(x,y)\nonumber\\
&&\!\exp(\!-j2k(\!x\sin\theta\cos\phi\!+\!y\sin\theta\sin\phi\!+\!z_0\cos\theta\!))d_xd_y
\end{eqnarray}
\setlength{\arraycolsep}{5pt}
simplifying (\ref{target_resp}) and using Fourier transform definition:
\begin{IEEEeqnarray}{rCl}
s(\theta,\phi)&=&\frac{exp(-2jk\cos\theta z_0)}{k\cos\theta}\nonumber\\
&&{\rm F_{2D}}(2k\sin\theta\cos\phi, 2k\sin\theta\sin\phi)
\end{IEEEeqnarray}
\begin{IEEEeqnarray}{rCl}
\label {recon_eq}
f(x,y)&={\rm F_{2D}}^{-1}((\exp(2jkz_0\cos\theta )(k\cos\theta)s(\theta,\phi)).
\end{IEEEeqnarray}
Here, ${\rm F_{2D}}$ indicates two dimensional Fourier transform and $exp(2jkz_0\cos\theta )$ represents a phase correction factor. The phase correction required to compensate phase difference due to placement of a target at particular distance $z_0$ away from the antenna plane. Reconstructed target image can be focused at another distance $z_f$ by evaluating the phase correction factor at that particular distance. The received wave data is also multiplied with amplitude correction of $k\cos\theta$. This operation is similar to windowing data in wavenumber domain before applying two dimensional inverse Fourier transform equation to recover the target's reflectivity function.
$(k_x, k_y)=(2k\sin\theta\cos\phi, 2k\sin\theta\sin\phi)$ represent spacing in the wavenumber domain. $(k_x, k_y)$ should have uniform spacing before applying two dimensional inverse Fourier equation. $(\theta,\phi)$ can be chosen such that $(k_x,k_y)$ have uniform spacing in wavenumber domain:
\begin{equation}
\label{theta}
\theta=\arcsin\left( \frac{\sqrt{k_x^2+k_y^2}}{2k}\right)
\end{equation}
\begin{equation}
\label{phi}
\phi=\arctan\left(\frac{k_y}{k_x}\right)
\end{equation}
this eliminates the need for interpolation of data from polar scale to uniform scale in the wavenumber domain.

\section{System Parameters}
\label{sym_para}
Resolution of the reconstructed image formed using inverse Fourier transform depends on range of spatial frequencies used. Spatial frequencies $(k_x, k_y)$ vary from $-2k$ to $2k$ for the plane waves. This is true for infinite aperture size. However, for the finite aperture range of the frequencies seen by the antenna array, can be calculated by considering instantaneous spatial frequency of the PM wave (i.e the term $exp(-j(k_ux+k_vy+\sqrt{k^2-k_u^2-k_v^2}z_0))$ in (\ref{tot_rx})) arriving at the antenna array \cite{soumekh}:
\begin{IEEEeqnarray}{rCl}
\frac{\partial }{\partial k_u}((k_ux+k_vy+\sqrt{k^2-k_u^2-k_v^2}z_0))=\nonumber\\
x+\frac{-k_u}{\sqrt{k^2-k_u^2-k_v^2}}z_0
\end{IEEEeqnarray}
\begin{IEEEeqnarray}{rCl}
x+\frac{-k_u}{\sqrt{k^2-k_u^2-k_v^2}}z_0=x-z_0\tan\theta\cos\phi
\end{IEEEeqnarray}
this instantaneous spatial frequencies $k_u$ and $k_v$ are limited by aperture size $[-L,L]$. Thus,
\begin{IEEEeqnarray}{rCl}
-L\leq x-z_0\tan\theta\cos\phi\leq L
\end{IEEEeqnarray}
\begin{IEEEeqnarray}{rCl}
\frac{x-L}{z_0}\leq\tan\theta\cos\phi\leq\frac{x+L}{z_0}.
\end{IEEEeqnarray}
On the similar lines we can show that,
\begin{IEEEeqnarray}{rCl}
\frac{y-L}{z_0}\leq\tan\theta\sin\phi\leq\frac{y+L}{z_0}.
\end{IEEEeqnarray}
Thus, squaring and adding above two inequalities,
\begin{IEEEeqnarray}{rCl}
\tan^2\theta\leq\min\left({\lvert\frac{y+L}{z_0}\rvert},{\lvert\frac{y-L}{z_0}}\rvert\right)^2+\nonumber\\\min\left({\lvert\frac{x+L}{z_0}\rvert},{\lvert\frac{x-L}{z_0}}\rvert\right)^2.
\end{IEEEeqnarray}
Range of $(\theta,\phi)$ found from the above inequalities, decide the range of frequencies $(k_x,k_y)$ which in turn affect the resolution of the reconstructed image. It is difficult to find close form expression for the resolution. However, it can be easily seen that, the resolution vary with the target location $(x,y,z_0)$. For a target situated in the plane at a distance $z_0$ from the antenna array, maximum resolution is obtained at the origin $(0,0,z_0)$. As the target moves away from this location, its resolution decreases. Also, as length of the aperture increases resolution of the system increases. Maximum resolution obtained by imaging system is limited by the Rayleigh limit $\lambda/2$ where $\lambda$ is the wavelength of the imaging system.
Thus, cross range resolution in $X$ direction is given by,
\begin{equation}
\label{resx}
\delta_x=\frac{2\pi}{{\rm Range}(k_x)}\approx\frac{\lambda}{2} \frac{\sqrt{x^2+y^2+z_0^2}}{ ({N_a}-1)\times \triangle_{xy}} \frac{1}{\lvert\cos\theta\lvert}
\end{equation}
where $\theta$ is the elevation look up direction corresponding to the target coordinates $(x,y,z_0)$. Same equation can be derived for the cross range resolution in $Y$ direction $\delta_y$. Detail comments about resolution equation are made in section \ref{sect_switched}.

\section{Review of Switched Array Method}
\label{sect_switched}
In this method\cite{sheen}, the reflected wave data is collected by putting only one transmitting and one receiving antenna in action at a given time. Assume that transmitting and receiving antenna are located very close to each other, then they can be represented by the mid-point between them, which is denoted by coordinates $(x^{\prime},y^{\prime})$ in Fig.~{\ref{fig_switch}}. Rest of the geometry remains same as explained earlier sections, then the total received reflected wave data from the target is given by,
\begin{IEEEeqnarray}{rCl}
\label{s}
s(x^\prime,y^\prime)&=&\int\int\exp(-j2k\sqrt{(x^\prime-x)^2+(y^\prime-y)^2+z_0^2})\nonumber
\\ &&f(x,y)dxdy
\end{IEEEeqnarray}
inversion equations is applied to $s(x^\prime,y^\prime)$ to recover $f(x,y)$  in \cite{sheen} ignore amplitude correction factor \cite{silverstein}. Following equation gives reconstruction with both the amplitude and phase correction factor. Dropping distinction between primed and unprimed coordinate system as they coincide the reconstruction equation is given by,
 \begin{IEEEeqnarray}{rCl}
\label{recon_sheen_corr}
f(x,y)&=&\rm FT^{-1}_{2D}(\rm FT_{2D}[\mit s(x,y)]\frac{1}{k_z}\exp(-jk_zz_0))
\end{IEEEeqnarray}
where,
 \begin{IEEEeqnarray}{rCl}
\label{FTdef}
{\rm FT_{2D}}[s(x,y)]&\!=\!&\!\!\int\!\!\int\! \!\!s(x,y)\!\exp\!(\!-\!j(k_xx\!+\!k_yy))dk_xdk_y.
\end{IEEEeqnarray}
Fourier transform variable $k_x$ and $k_y$ range from $-2k$ to $2k$ and satisfy $k_x^2+k_y^2+k_z^2=(2k)^2$. In (\ref{FTdef}) range of $k_x$ and $k_y$ over which $k_x^2+k_y^2\leq (2k)^2$ is known as visible region over which two dimensional spatial Fourier transform is calculated. Inbuilt Fourier transform functions in MATLAB do not consider this visible region condition and hence cannot be directly used for the image reconstruction.

We propose the new reconstruction method based on matched filtering approach, which incorporates the amplitude correction factor as well as effect of the finite aperture size. In addition, this method is more suitable for implementation using MATLAB.

For a point target placed at the origin of target's coordinate system $(0,0,z_0)$ the received reflected wave data is  given by,
\begin{IEEEeqnarray}{rCl}
\label{h}
h(x^\prime,y^\prime)&=&\exp(-j2k\sqrt{(x^\prime)^2+(y^\prime)^2+z_0^2})
\end{IEEEeqnarray}
$h(x^\prime,y^\prime)$ in (\ref{h}) is the impulse response or the point spread function of the imaging system. Dropping distinction between primed and unprimed coordinate system as they coincide, (\ref{s}) can be rewritten as,
\begin{IEEEeqnarray}{rCl}
\label{sconv}
s(x,y)&=&f(x,y)\star h(x,y).
\end{IEEEeqnarray}
Thus, target's reflectivity function can be recovered by following matched filtering operation:
\begin{IEEEeqnarray}{rCl}
\label{mf}
f(x,y)&=&\rm F_{2D}^{-1}[(\rm F_{2D}\mit (s(x,y))(\rm F_{2D}\mit(h_{z_f}(x,y))^\star)
\end{IEEEeqnarray}
$h_{z_f}(x,y)$ is the impulse response evaluated at the focussing distance $z_f$ instead of $ z_0$ Here, $\rm F_{2D}$ defines two dimensional Fourier transform in the following manner:
 \begin{IEEEeqnarray}{rCl}
\label{FTm}
{\rm F_{2D}}[s(x,y)]&=&\!\int\!\!\int\!\! s(x,y)\!\exp\!(-j(\omega _1x\!+\!\omega_2y))dxdy.
\end{IEEEeqnarray}
 As transform defined in (\ref{FTm}) does not consider any visibility condition like (\ref{FTdef}), inbuilt MATLAB functions such as FFT2 can be used for the reconstruction. The resolution of reconstred image is given by \cite{sheen}:
\begin{equation}
\label{resswitched}
\delta_x=\frac{\lambda}{2}\frac{R}{D}\approx \frac{\lambda}{2}\frac{z_0}{( {N_a}-1)\times \triangle_{xy}}
\end{equation}
where, $R$ is the range and $D$ is the size of aperture. When $R\gg D$ we can approximate, $R\approx  z_0$ and resolution can be obtained in terms of system parameters.
For a given $z_0$, the resolution increases as the number of antenna elements increase. This argument is consistent with (\ref{mf}). It can be seen that the resolution depends on correlation between $h(x,y)$ and $\rm conj\mit(h_{z_f}(x,y))$. For an infinite aperture size, this correlation is impulse function in spatial domain and thus gives the maximum resolution. With reduction in the aperture size, the resolution decreases, as the correlation is no longer an impulse function.

The resolution is also a function of distance between target and antenna array i.e. $z_0$. Bringing the target closer to certain extent increases the resolution. However, after some distance if the target is brought further closer, then the antenna elements near edge of the aperture receive aliased signal. Same argument is true for increasing the spacing between antenna elements ($\triangle_{xy}$) keeping the other parameter constant. As shown in \cite{sheen}, for the system in which $z_0>>$ size of aperture the antenna spacing should be at least $\lambda/2$ to avoid aliasing.  Increasing $N_a$ and keeping the other parameters constant has the higher limit in terms of $N_a$, before aperture size becomes comparable with $z_0$, which results into aliasing.

Table \ref{table:switch} summarizes the MATLAB implementation of switched array method reconstruction algorithm for two dimensional target.
\begin{table}[ht]
\caption{Summary of Reconstruction using switched array method}
\centering
\begin{tabular}{c l}
\hline
1&  Collect $s(x,y )$ by the transceiver operation as shown in (\ref{s}).\\
2 & Calculate the impulse response $h_{z_f}(x,y)$ as per (\ref{h}).\\
3 & Compute Fourier transforms using inbuilt MATLAB commands.\\
4 & Reconstruct using the equation (\ref{mf}).
\end{tabular}
\label{table:switch}
\end{table}

To compare the beam steered array imaging system and the switched array imaging system, their performance is considered in the presence of noise. Noise at the each receiving antenna is assumed to be i.i.d. Gaussian with zero mean and finite variance.

For the switched array system, only one antenna illuminates the target and collects reflection at a given time. Whereas, for the two dimensional beam steering imaging system, the whole antenna array is in operation. Thus, the target is illuminated using powerful directional beam. In addition, the receiver coherently combines only the waveforms coming from a particular direction.

Thus, the two dimensional beam steering method offers array gain in terms of SNR over the switched array method in both transmit and receive direction. Amount of SNR gain obtain increases with the aperture size. It is $N_a \times N_a$ in both the transmit and receive mode \cite{goldsmith}.

\section{Performance in the Presence of Phase Noise}
\label{phasenoise}

Phase shifts are introduced at each antenna before transmitting and after receiving the data to steer the beam in the particular direction. These phase shifters have finite resolution depending on number of bits of ADC used for the quantization of the phase.

For example, one bit ADC at each antenna would either transmit the waveform without any phase shift or invert and transmit it ($180^0$ Phase shift). The beam steering based reconstruction can be carried out with different number of quantization bits.

This phase quantization impacts the array gain obtained in the presence of additive Gaussian noise which is discussed in earlier section \cite{Bakr:EECS-2009-1}. However, the loss in the gain is for the worst case, i.e. for one bit phase quantizer is limited to $-3.9$ dB.

Quantizing phases of the beam steering weights add phasenoise and affect the resolution of the reconstructed target. Apart from the quantization of beam steering weights other factors such as timing jitter while sampling the waveforms add to uncertainty in the phase at each transmitting antenna. Thus, the resulting beam steering suffers from deformations such as tilt and increase in sidelobes. From the observations, as the variance of phase noise increases, first sidelobes grow higher and then resulting beam oscillate around intended direction giving rise to tilt in the resulting pattern. As the sidelobes compete with the main lobe, the received wave data is corrupted with data from the undesired direction. Thus, the reconstruction of the target is not faithful.

These impacts of the phase noise on target image reconstruction can be analyzed using "Point Spread Function (PSF)". Which is the response of the imaging system to the point target placed at origin of target's coordinate system. To obtain PSF, we consider one dimensional version of two dimensional beam steering imaging system discussed earlier. Antenna array with transmitter location $(u, 0)$ and receiver location $(a,0)$ is used. Target is situated at distance $z=z_0$ in $X-Z$ plane. For an imaging system, steering beam at angle $\theta$, response of point target at situated at $(x,z_0)$ is given by,

\begin{IEEEeqnarray}{rCl}
\label{Rxsig1d}
s_{x}(\theta)&=&\int exp(-jk\sqrt{(x-u)^2+z_0^2})W_{\theta}(u)\nonumber
\\ &&\left[\int exp(-jk\sqrt{(x-a)^2+z_0^2})W_{\theta}(a)d_a\right]d_u
\end{IEEEeqnarray}
where $W_{\theta}(u)=\exp(-jku\sin\theta)$ is the beam steering weight. Associated with the weight, coresponding to each antenna element in transmit direction, consider a the phase noise $\psi_u$, which is an additive i.i.d Gaussian with zero mean and standard deviation $\sigma_{\phi}$. For example, noisy weight at the each antenna in transmit direction is given by (\ref{eqn_nweights}). This phase noise is added to the actual weights in both transmit $W_{\theta}(u)$ and receive direction $W_{\theta}(a)$.
\begin{equation}
\label{eqn_nweights}
\hat W_{\theta\phi}(u)=\exp(-j(ku\sin\theta+\psi_u ).
\end{equation}
$\hat s_{x}(\theta)$ is the received reflected wave data, obtained using these noisy weights:
\begin{IEEEeqnarray}{rCl}
\label{Rxsig1d}
\hat s_{x}(\theta)&=&\int exp(-jk\sqrt{(x-u)^2+z_0^2})\hat W_{\theta}(u)\nonumber
\\ &&\left[\int exp(-jk\sqrt{(x-a)^2+z_0^2})\hat W_{\theta}(a)d_a\right]d_u.
\end{IEEEeqnarray}
Following is the reconstruction equation in the presence of phase noise:
\begin{IEEEeqnarray}{rCl}
\label {recon_eq1d}
\hat f(x)&={\rm F}_{\rm 1D}^{-1}((\exp(2jkz_0\cos\theta )(k\cos\theta)\hat s_x(\theta)
\end{IEEEeqnarray}
here, $\rm F^{-1}_{ 1D}$ is one dimensional inverse Fourier transform. For a point target situated at $(0, z_0)$ shape of ideal reconstructed image (PSF) would be an impulse situated at the origin. However, in the practice Sinc like shape is obtained due to the finite aperture size. Thus, PSF shows sidelobes, which along with main lobe indicate directional selectivity of the beam steering system. If sidelobe has higher amplitude, reflected wave data is ambiguously collected from many other directions than the desired one. Hence, sidelobe's amplitude should be as low as possible as compared to main lobe.  A metric called sidelobe suppression level (SSL), which is ratio of side lobe to main lobe is computed to compare the resolution of the reconstructed images.

For an imaging system operating at the frequency $f_c$, timing jitter of $\triangle_t$ corresponds to $\sigma_{\phi}= 2\pi f_c \triangle_t$. If the imaging system is operating at $60$ GHz, $3$ ps of timing jitter leads to  $\sigma_{\phi}= 1.13$ radians. We can obtain the expected PSF for the different values $\sigma_{\phi}$. This analysis is done for different number of antenna elements. The spacing between antenna elements is varied to keep the total aperture size $N_a\times \triangle_{x}$ constant. Aperture size is kept constant to obtain focussed reconstruction of the target kept at fixed distance $z_0.$

The SSL remains constant for the low values of $\sigma_{\phi}$ upto a certain threshold after which it increases linearly with the $\sigma_{\phi}$. This threshold is defined as sidelobe suppression breakpoint $\sigma_{\phi}^{\rm SB}$ \cite{accardi}. Once the phase noise increases such that $\sigma_{\phi}>\sigma_{\phi}^{\rm SB}$, rise in the sidelobes cannot be prevented using the time averaging. In other words the imaging system is unstable in the region where $\sigma_{\phi}>\sigma_{\phi}^{\rm SB}$. The $\sigma_{\phi}^{\rm SB}$ increases with increase in the aperture size, indicating that the large arrays are more immune to the phase noise. Thus, given the number of antennas used we can numerically find tolerable phase noise that imaging system can handle without being unstable.

\begin{figure}
\centering
\includegraphics[width=3.5in]{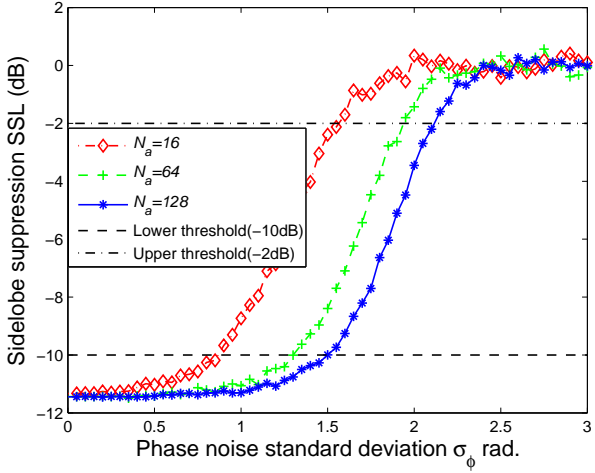}
\caption{Sidelobe suppression against phase noise standard deviation	}
\label{SSL1}
\end{figure}

\begin{figure}[!tl]
\centering
\includegraphics[width=3.5in]{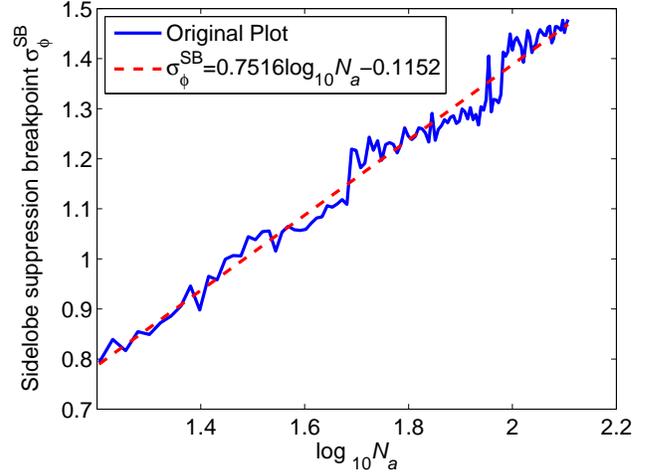}
\caption{Sidelobe suppression breakpoint against number of antenna elements }
\label{PhiSB}
\end{figure}

Number of antenna elements are varied from $16$ to $128$ and SSL is calculated for given number of antenna elements with $\sigma_{\phi}$ varying from $0$ to $5$ radians. The results are shown in Fig.~{\ref{SSL1}}. It shows variation of SSL against phase noise standard deviation for three different number of antenna elements $16, 64$ and $128$. The region where SSL lies between $-10$ dB to $-2$ dB is interpolated to obtain straight line equation. Lowest point of this line indicates the $\sigma_{\phi}^{\rm SB}$. Relation is obtained between the number of antenna elements and $\sigma_{\phi}^{\rm SB}$ by curve fitting as shown in Fig.~{\ref{PhiSB}}.
\begin{equation}
\label{resswitched}
\sigma_{\phi}^{\rm SB}=0.7516\log_{10}(N_a) - 0.1152.
\end{equation}

If the phase noise is increased above $\sigma_{\phi}^{\rm SB}$, then the sidelobes increase linearly with increase in phase noise. After certain threshold they become significant with respect to the main beam. Using the same interpolation techniques used for $\sigma_{\phi}^{\rm SB}$ we can find the upper threshold point $\sigma_{\phi}^{\rm TB}$ on phase noise so that SSL remains below -1 dB.
\begin{equation}
\label{resswitched}
\sigma_{\phi}^{\rm TB}=0.5326 \log_{10}(N_a) +1.1282.
\end{equation}

\begin{figure}
\centering
\includegraphics[width=3.5in]{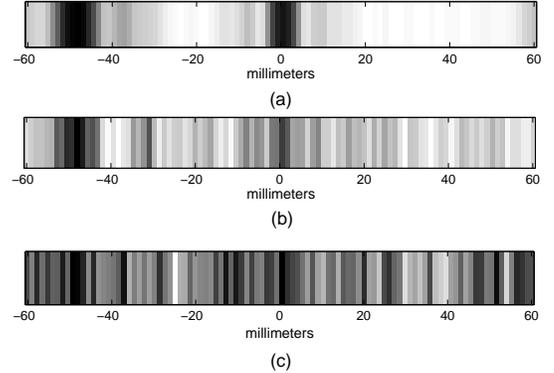}
\caption{Reconstruction of two reflectors placed at distance $100$ mm from an antenna array :(a) $\sigma_{\phi}=\sigma_{\phi}^{\rm SB}$; (b) $\sigma_{\phi}^{\rm SB}< \sigma_{\phi} < \sigma_{\phi}^{\rm TB}$; (c) $\sigma_{\phi} > \sigma_{\phi}^{\rm TB}$ }
\label{phasen}
\end{figure}

As the phase noise grows above the $\sigma_{\phi}^{\rm TB}$, the mainlobe and the sidelobe are hardly distinguishable from each other. In this case, the transmitted beam would also illuminate the objects in undesired direction also the receiver beam steering would receive the reflected wave data from unwanted direction. Thus, the reconstruction of the target obtained when phase noise is above $\sigma_{\phi}^{\rm TB}$ is unfaithful.

To illustrate the above result consider $16$ elements one dimensional imaging system operating at the frequency of $60$ GHz. The antenna elements are located on $X$ axis and two target reflectors are located $X-Z$ plane at $(50,100)$ and $(0,100)$. Off-broadside target reflects $3$ dB more than the broadside target. Using the above equations $ \sigma _ {\phi} ^ {\rm SB} =0.7898$ radians and $\sigma_{\phi}^{\rm TB}=1.7695$ radians. The reconstruction results are shown in Fig.~{\ref{phasen}}. For $\sigma _ { \phi}=\sigma _{ \phi}^{\rm SB} $ the reconstruction is still possible. However, when $\sigma_{\phi}$ increases above $\sigma_{\phi}^{\rm SB}$ imaging system becomes unstable, loses directional selectivity and broadside target is no longer detected. Finally, when $\sigma_{\phi} $ goes above $\sigma_{\phi}^{\rm TB}$, the target cannot be reconstructed faithfully.

\section{Simulation Results}
\label{symres}

\begin{figure}
\centering
\includegraphics[width=3.5in]{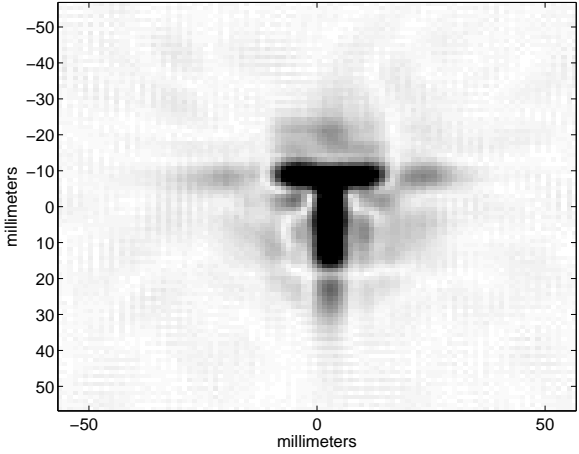}
\caption{Reconstructed image with $32\times 32$ array (beam steering method SNR=$-50$ dB) }
\label{beamnoise}
\end{figure}

\begin{figure}
\centering
\includegraphics[width=3.5in]{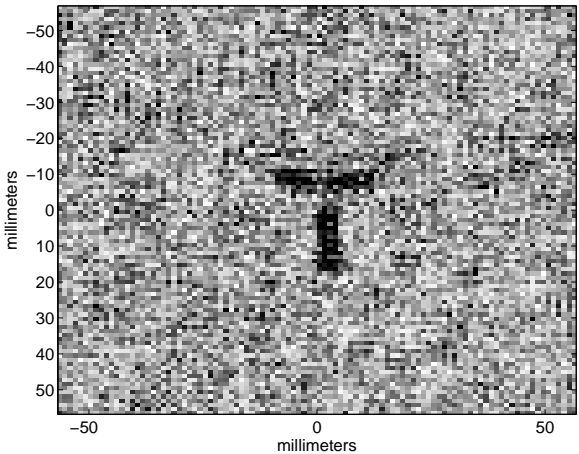}
\caption{Reconstructed image with $32\times 32$ array (switched array method SNR=$-50$ dB) }
\label{switchednoise}
\end{figure}

Simulations are carried out using millimeter wave of frequency $60$ GHz. 'T' shaped target is used. It has height of $27.5$ mm and width of $22.5$ mm. The target is placed at distance $100$ mm from the antenna plane. $32\times 32$ antenna array with $1.75$ mm spacing between antenna elements is in operation for the reconstruction. At each receiving antenna additive i.i.d Gaussian noise is added such that SNR$=-50$ dB. Steering angles are obtained so as to have uniform spacing in wavenumber domain. To analyze performance in the presence of additive white Gaussian noise, the phase shifters in this operation are assumed to have infinite resolution. The reconstructions is carried out using the two dimensional beam steering algorithm and the result is shown in Fig.~{\ref{beamnoise}}.

To compare performance of the two dimensional beam steering method with the switched array method, parameters such as system geometry ($N_a$ and $\triangle_{xy}$), target shape, reflectivity and its distance from the array ($z_0$) are kept same. In addition, the average signal power transmitted is kept same for the analysis. The total transmitted power from the each antenna element in switched array method is same as sum of the all antenna element's power in the two dimensional beam steering case. Also, additive Gaussian noise is added at each receiving antenna in switched array case is assumed to i.i.d. With the same SNR=$-50$ dB reconstruction is carried out using the switched array method. Fig.~{\ref{switchednoise}} shows the result. It can be seen that the two dimensional beam steering method performs better than the switched array method. This improvement is achieved due to the beamforming array gain of $60$ dB.

\begin{figure}
\centering
\includegraphics[width=3.5in]{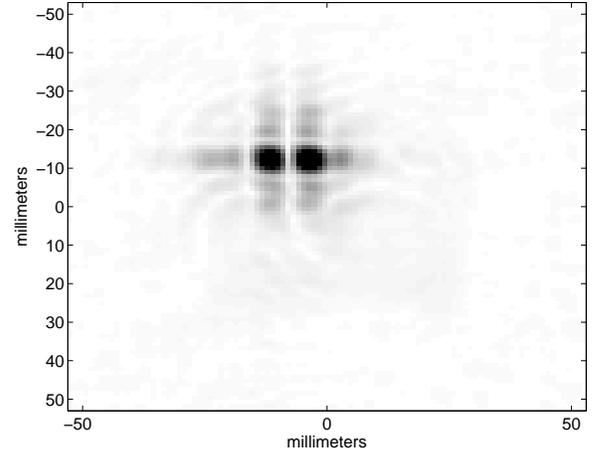}
\caption{Reconstruction  of two reflectors 10 mm apart using two bits quantization (off broadside) }
\label{2bitsoffbroadside}
\end{figure}

\begin{figure}
\centering
\includegraphics[width=3.5in]{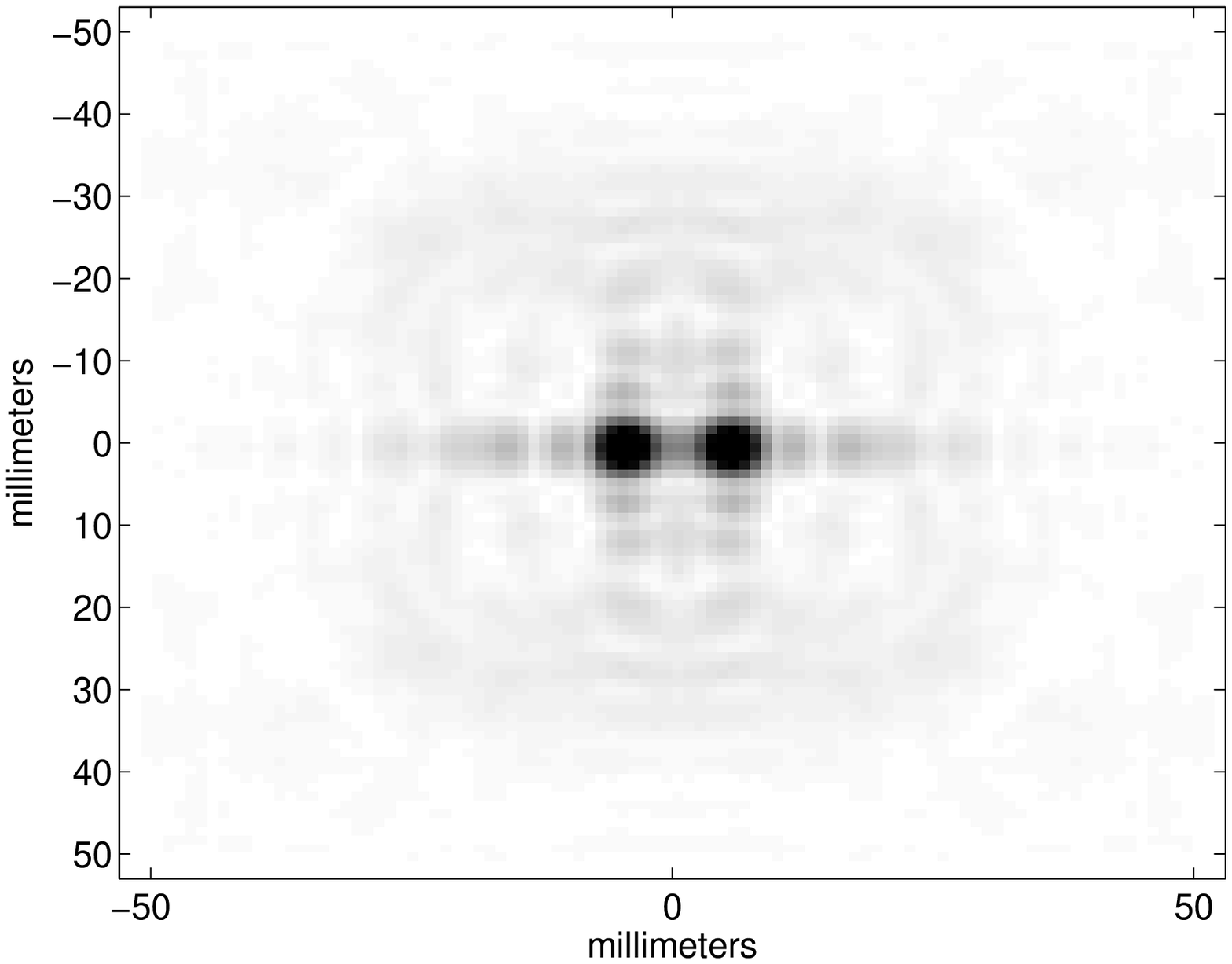}
\caption{Reconstruction of two reflectors 10 mm apart using two bits quantization (broadside) }
\label{2bitsbroadside}
\end{figure}

\begin{figure*}
\centering
\includegraphics[scale=0.42]{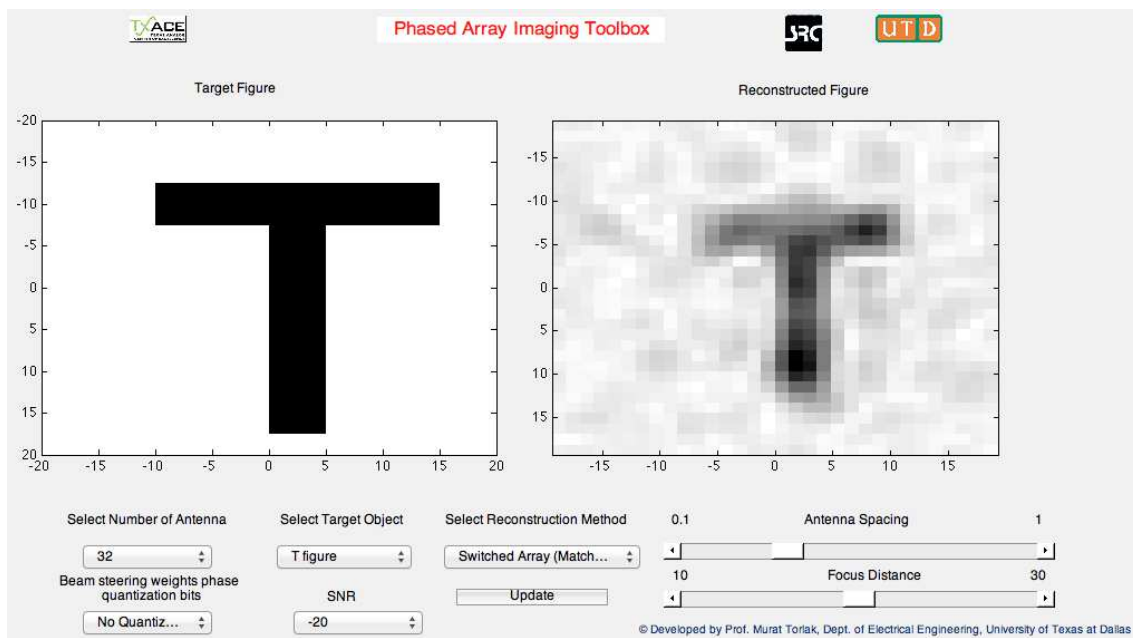}
\caption{Phase array Imaging toolbox showing reconstruction in the presence of noise}
\label{toolbox}
\end{figure*}

Fig.~{\ref{2bitsoffbroadside}} and {\ref{2bitsbroadside}}  demonstrate shift varying resolution of the imaging system. Here, two reflectors are placed at a distance $10$ mm from each other and $100$ mm from the antenna array. The two dimensional beam steering method is used for reconstruction, with the same setup as described in beginning of the section. However, $2$-bit phase quantizer is used at the each phase shifters in transmit and receive direction. The reconstructions of the target placed at off-broadside and broadside positions with respect to antenna array are shown in Fig.~{\ref{2bitsoffbroadside}} and {\ref{2bitsbroadside}} respectively.  Fig.~{\ref{2bitsoffbroadside}} shows the two reflectors being reconstructed at the closer distance with respect to each other than the Fig.~{\ref{2bitsbroadside}}. Therefore, the resolution of system degrades as the target is placed at the location other than broadside.

A phase array imaging toolbox was developed by the authors which demonstrates different imaging method discussed in this paper \cite{toolbox}. Fig.~{\ref{toolbox}} shows a snapshot of the toolbox, which uses switched array method of reconstruction with matched filtering technique. The reconstruction is carried out in the presence of noise with SNR=$-20$ dB. This toolbox enables user to model end-to-end phase array imaging system. Various system parameters such as array size, antenna spacing and focus distance can be modified and their effect on the image reconstruction can be studied. In addition, RF impairments such as quantization of the beam steering weights and noise can be added to the ideal reconstruction. Future work includes expanding the toolbox to include various antenna patterns and coupling between antenna elements \cite{odzemir2}, \cite{odzemir3}.\nocite{soumekhbookII}\nocite{soumekhpaperII}

\section{Conclusions}

In this paper we derive two dimensional beam steering algorithm for imaging applications. The beam steering angles are obtained in such a way that no interpolation technique is required in either wavenumber or spatial domain. This two dimensional beam steering method of reconstruction has better performance in the presence of noise than the switched array method. We also reviewed the switched array method and developed the reconstruction framework using matched filtering, which is suitable for MATLAB implementation. For comparison of different millimeter wave imaging system, a MATLAB based toolbox is constructed. A numerical analysis is performed to relate aperture size with the maximum phase noise variance that imaging system can tolerate.\nocite{soumekhbookI}\nocite{soumekhpaperIII}\nocite{sheenpaperII}

\bibliographystyle{IEEEtran}
\bibliography{beampaper}

\end{document}